\documentclass[conference]{IEEEtran}

\usepackage{cite}
\usepackage{amsmath,amssymb,amsfonts}
\usepackage{algorithmic}
\usepackage{graphicx}
\usepackage{textcomp}
\usepackage[dvipsnames,table]{xcolor}
\usepackage{amsmath}
\usepackage{makecell}
\usepackage{booktabs}
\def\BibTeX{{\rm B\kern-.05em{\sc i\kern-.025em b}\kern-.08em
    T\kern-.1667em\lower.7ex\hbox{E}\kern-.125emX}}
\begin{document}

\title{Recognition of Dysarthria in Amyotrophic Lateral Sclerosis patients using Hypernetworks}

\author{\IEEEauthorblockN{Loukas Ilias, Dimitris Askounis}
\IEEEauthorblockA{\textit{DSS Lab, School of ECE, NTUA} \\
\textit{15780 Athens, Greece} \\
\{lilias,askous\}@epu.ntua.gr}
}

\maketitle

\begin{abstract}
Amyotrophic Lateral Sclerosis (ALS) constitutes a progressive neurodegenerative disease with varying symptoms, including decline in speech intelligibility. Existing studies, which recognize dysarthria in ALS patients by predicting the clinical standard ALSFRS-R, rely on feature extraction strategies and the design of customized convolutional neural networks followed by dense layers. However, recent studies have shown that neural networks adopting the logic of input-conditional computations enjoy a series of benefits, including faster training, better performance, and flexibility. To resolve these issues, we present the first study incorporating hypernetworks for recognizing dysarthria. Specifically, we use audio files, convert them into log-Mel spectrogram, delta, and delta-delta, and pass the resulting image through a pretrained modified AlexNet model. Finally, we use a hypernetwork, which generates weights for a target network. Experiments are conducted on a newly collected publicly available dataset, namely VOC-ALS. Results showed that the proposed approach reaches Accuracy up to 82.66\% outperforming strong baselines, including multimodal fusion methods, while findings from an ablation study demonstrated the effectiveness of the introduced methodology. Overall, our approach incorporating hypernetworks obtains valuable advantages over state-of-the-art results in terms of generalization ability, parameter efficiency, and robustness.
\end{abstract}


\section{Introduction}

Amyotrophic lateral sclerosis (ALS) is a progressive neurodegenerative disease, which damages both upper and lower motor neurons over time. This degeneration results in dysphagia, impairment of speech intelligibility due to dysarthria, limb paralysis, and respiratory failure \cite{joubert2011speech,Chi01012009,al2013epidemiology,xu2020global}. Studies have shown that ALS affects between 4.1 and 8.4 per 100,000 persons \cite{longinetti2019epidemiology}. Although the survival time is approximately three years, approximately 20\% of individuals with ALS live for five years, 10\% survive for 10 years, and 5\% live for 20 years or more \cite{als_stages}. Diagnosis of ALS is a difficult task \cite{paganoni2014diagnostic}, since there is no single medical test. However, early diagnosis is crucial, so as to ensure that individuals will receive treatment to slow ALS progression and maintain a good Quality of Life \cite{chio2004cross, als_diagnosis}. Since ALS leads to dysarthria, researchers have proposed methods for recognizing dysarthria in ALS patients. The recognition of dysarthria is achieved by speech-language pathologists through the Revised ALS Functional Rating Scale (ALSFRS-R) \cite{CEDARBAUM199913}. Specifically, ALSFRS-R indicates severity levels of dysarthria from 0 to 4, where 0 indicates loss of useful speech, 1 denotes speech combined with nonvocal communications, 2 indicates intelligible with repeating, 3 denotes detectable speech disturbance, and 4 denotes normal speech. 

Existing studies rely on feature extraction strategies followed by feature selection techniques and train of traditional Machine Learning (ML) algorithms. However, this is a tedious procedure demanding feature expertise, while it is not ensured that the optimal set of features is found. Recently, researchers extract log-Mel spectrograms or Mel-frequency Cepstral Coefficients (MFCC), their delta, and double-delta and train deep neural networks. However, these approaches use customized Convolutional Neural Networks (CNNs), obtaining often suboptimal performance, since these approaches depend on limited data. Also, these customized CNNs are followed by dense layers, which are accountable for doing everything and thus do not adopt an input-conditional computation logic. On the other hand, hypernetworks \cite{ha2017hypernetworks, chauhan2024brief} constitute a powerful deep learning technique, which ensures greater flexibility adaptability, faster training, information sharing, model compression, and so on. Hypernetworks have been proved advantageous for multiple tasks, including causal inference, natural language processing, transfer learning, and weight pruning. These networks are also beneficial for tasks with limited data. Specifically, hypernetworks are neural networks, which generate weights for another neural network, namely the target network. By generating weights dynamically, hypernetworks introduce implicit regularization, improving generalization performance.

To tackle the aforementioned limitations, we present the first study for recognizing dysarthria in ALS patients using hypernetworks. Specifically, we use audio files, which correspond to syllable repetitions of /pa/, and transform them into images of three channels, namely log-Mel spectrogram, delta, and delta-delta. Next, we pass each image through an AlexNet pretrained model and get an image representation vector. After that, motivated by the fact that hypernetworks ensure faster training and are beneficial for tasks with limited data, we employ a hypernetwork, which receives as input a condition vector, which follows normal distribution. This hypernetwork generates weights for the target network, which receives as input the output of AlexNet. Finally, the output layer corresponds to the prediction of dysarthria, i.e., binary classification task. Results are performed on the publicly available VOC-ALS dataset \cite{dubbioso2024voice}. Finally, a series of ablation experiments is performed for exploring the effectiveness of the introduced method. Results show that our proposed approach improves state-of-the-art ones, while also offering multiple advantages over existing research initiatives due to the inherent benefits of hypernetworks, i.e., parameter efficiency, task adaptation - robustness, generalization ability.

Our main contributions can be summarized as follows:
\begin{itemize}
    \item To the best of our knowledge, this is the first study using hypernetworks to detect dysarthria in ALS patients.
    \item We compare our approach with strong baselines, including multimodal fusion methods.
    \item We perform a series of ablation experiments to explore the effectiveness of our approach.
\end{itemize}

\section{Related Work} \label{related_work}

\subsection{Traditional Machine Learning Algorithms} \label{traditional_ML}

Dubbioso et al. \cite{DUBBIOSO2024105706} extracted a set of acoustic features from different tasks, performed feature selection strategies, and trained a Decision Tree classifier for differentiating healthy subjects from non-healthy ones and predicting dysarthria severity levels in ALS patients. Experiments were performed on different tasks, including reading, monologue, and vocalization. In \cite{wisler-etal-2019-speech}, the authors extracted a set of acoustic and articulatory features and trained Ridge regression and a Support Vector Machine to predict the ALSFRS-R score. In terms of the acoustic features, the authors used MFCC, their delta, and delta-delta and computed some statistics, e.g., mean, standard deviation. Regarding articulatory features, the authors computed a distance matrix and computed some statistics, e.g., skewness, kurtosis, and so on. The study in \cite{info:doi/10.2196/21331} was focused on the detection of ALS with bulbar involvement. The authors extracted a set of features, including jitter, shimmer, harmonics-to-noise ratio, pitch, and so on. Principal component analysis was used for dimensionality reduction. Finally, the authors trained the following machine learning classifiers: SVM, neural network with a hidden layer, LDA, LR, Naive Bayes, and Random Forest (RF). Vashkevich and Rushkevich \cite{VASHKEVICH2021102350} proposed a study based on voice analysis to detect ALS patients. Specifically, the authors extracted a set of acoustic features from phonation vowels /a/ and /i/, performed feature selection algorithms, and trained a linear discriminant analysis for the classification purposes. The authors in \cite{simmatis2024detecting} extracted a set of acoustic features and trained a bayesian logistic regression model for differentiating the following groups: \textit{(i)} control vs ALS, \textit{(ii)} control
vs ALS-early, and \textit{(iii)} ALS-early vs ALS-late . The main limitation of this study is related to the imbalanced dataset between ALS and control participants. Specifically, the dataset includes 119 ALS patients and 22 healthy controls.

\subsection{Deep Neural Networks} \label{DNNs_related}

Two different transfer learning strategies were introduced in \cite{bhattacharjee23_interspeech}. Specifically, the authors explored fine-tuning and multitask learning frameworks. As auxiliary tasks, the authors used input feature reconstruction and gender classification. The authors used as input to the deep neural networks a vector consisting of MFCC (excluding energy coefficient) with delta and double delta features. The deep neural network comprised a series of dense layers. Three set of experiments were performed in the study of \cite{9179503}, including \textit{(1)} classification among ALS, Parkinson disease (PD), and Healthy control, \textit{(2)} 5-class ALS severity classification based on ALSFRS-R, and \textit{(3)} 3-class PD severity classification. The authors used as input log-Mel spectrograms and passed them through CNN layers followed by fully connected layers. Four tasks were used, including spontaneous speech, image description, sustained phonation, and diadochokinetic rate. In \cite{vieira2022machine}, the authors segmented the audio file into non-overlapping audio frames, converted it into log-Mel spectrogram and passed each frame through CNN layers. Next, they aggregated each frame's output to get the final prediction for the entire voice signal. The task was the prediction of the ALSFRS-R score. Three approaches were employed for classifying ALS patients and healthy control in \cite{an18c_interspeech}. In terms of the first approach, the authors extracted features using the openSMILE toolkit and trained an Artificial Neural Network with one hidden layer. Regarding the other two approaches, the authors utilized filterbank, delta, and delta-delta as input to time-CNNs and frequency-CNNs.

\subsection{Related Work Review Findings}

As is evident in Section~\ref{traditional_ML}, existing studies focus on the extraction of acoustic features and the train of shallow machine learning classifiers, which constitutes a tedious procedure and does not generalize to new subjects. As is presented in Section~\ref{DNNs_related}, existing studies convert the audio files into log-Mel spectrograms, delta, and delta-delta and pass them through CNN layers followed by dense layers.

Our study is different from existing studies, since we present the first study incorporating hypernetworks into a deep neural network for recognizing dysarthria in ALS patients. Also, this study has been performed in a newly collected dataset, which is publicly available.

\section{Dataset and Task} \label{dataset_task}

We use the VOC-ALS dataset described in \cite{dubbioso2024voice} to perform our experiments. VOC-ALS is a newly collected publicly available dataset and represents the most comprehensive freely downloadable dataset. Specifically, this dataset comprises 51 healthy controls and 102 ALS patients. Each participant is asked to perform a series of tasks using a smartphone application, including phonation of the vowels /a/, /e/, /i/, /o/, /u/ and /pa/, /ta/, /ka/ syllable repetition, recordings of subjects vocalizing the days of the week, reading task, and monologue. However, the authors have made publicly available only the audio files corresponding to the phonation of the vowels /a/, /e/, /i/, /o/, /u/ and /pa/, /ta/, /ka/ syllable repetition.

In this study, the task is to categorize ALS patients into dysarthric and non-dysarthric ones. According to the ALSFRS-R score, four classes are available in terms of the ALS patients, including severe, moderate, mild, and non-dysarthric. A detailed description of the number of patients and demographic information per category is reported in Table~\ref{dataset_description}. For performing our experiments, we merge severe, moderate, and mild classes into one class, i.e., dysarthric.

\begin{table}[!htb]
\tiny
    \centering
    \caption{Description of VOC-ALS dataset.}
    \begin{tabular}{c|cccc}
    \toprule
    Severity class     &  Severe & Moderate & Mild & Normal \\ \midrule
    ALSFRS-R     &  1 & 2 & 3 & 4 \\
    \# M: \# F & 2:3 & 9:4 & 21:10 & 33:20 \\
    Age (M:F) & 64.00:65.67 & 65.44:72.00 & 65.48:60.90 & 60.15:61.55 \\
    Total & 5 & 13 & 31 & 53 \\ \bottomrule
    \end{tabular}
    \label{dataset_description}
\end{table}

\section{Methodology} \label{methodology}

\begin{figure*}[!htb]
    \centering
    \includegraphics[width=0.67\linewidth]{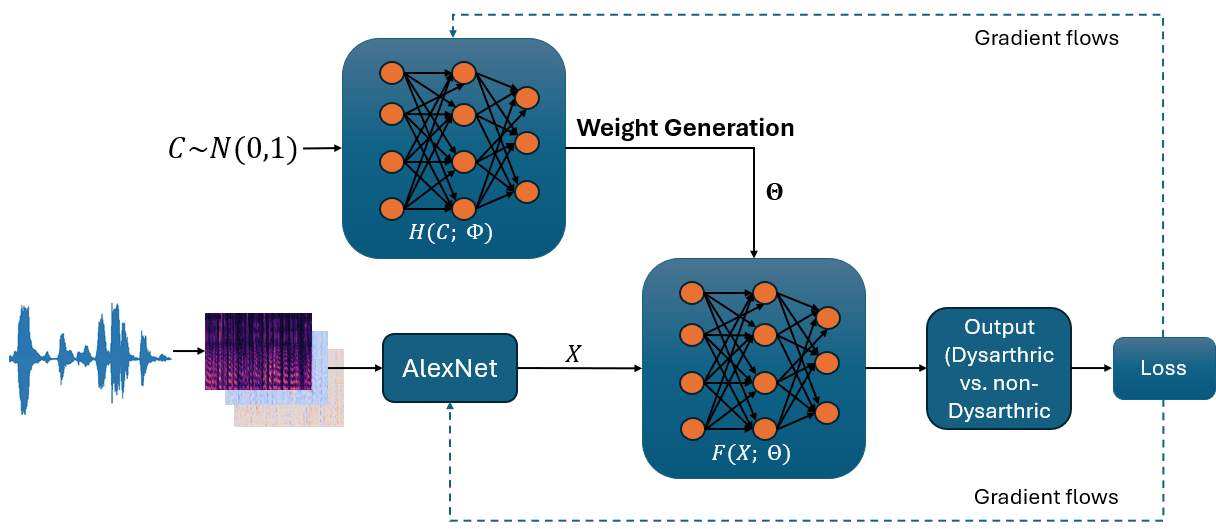}
    \caption{Illustration of our proposed methodology. Each speech signal is transformed into log-Mel spectrogram, delta, and delta-delta, and is given as input to a pretrained AlexNet model. The output vector of the AlexNet model with a dimensionality of 768 is given as input to a target network ($F(X; \Theta$), where its weights are generated by a hypernetwork ($H(C; \Phi)$). The input to the hypernetwork is denoted by $C$ and follows a normal distribution. Finally, we use an output layer consisting of two units, which differentiates dysarthric from non-dysarthric ALS patients. }
    \label{proposed_methodology}
\end{figure*}

In this section, we describe our introduced methodology for recognizing dysarthria in ALS patients. Below, we describe in detail each component of our proposed architecture. Our proposed approach is illustrated in Fig.~\ref{proposed_methodology}.

\noindent \textbf{Input.} We use repetitions of the syllable /pa/, denoted as \textit{rhythmPA} in the dataset. Next, we convert each audio file into log-Mel spectrogram, delta, and double-delta. To do this, we utilize the Python library, namely \textit{librosa} \cite{mcfee2015librosa}. In our experiments, we use 256 Mel bands, hop length accounting for 512. Let the input representation be $x \in \mathbb{R}^{3\times 224 \times 224}$.

\noindent \textbf{Deep Learning Model - AlexNet.} Next, we pass $x$ through a pretrained AlexNet model \cite{krizhevsky2014one}. AlexNet consists of five convolutional layers followed by fully connected layers. The first convolutional layer uses large filters (11x11) to capture low-level patterns, followed by progressively smaller filters ($5\times 5$ and $3\times 3$) to extract finer details. ReLU activation and max-pooling layers are used in AlexNet architecture. 

In our experiments, we modify AlexNet by removing the last dense (fully connected) layer. Let the output of the AlexNet model be $X \in \mathbb{R}^{D}$, where $D=768$.

\noindent \textbf{Hypernetwork.} Next, we use hypernetworks \cite{ha2017hypernetworks, chauhan2024brief}. Specifically, hypernetworks are neural networks which are trained for generating weights for another neural network, known as the target network. Unlike traditional models where parameters are directly optimized, a hypernetwork leverages a context vector $C$, which generates weights for the target network dynamically. The context vector $C$ can be task-, data-, or noise-conditioned, allowing the target network to adapt more efficiently. 

As illustrated in Fig.~\ref{proposed_methodology}, let the hypernetwork be denoted as $H(C; \Phi)$, which maps the context vector $C$ to the parameters $\Theta$ of the target network $F(X; \Theta)$. The output of the hypernetwork, $\Theta$, represents the generated weights and biases, which is given to the target network. In our experiments, we set $C$ to follow a normal distribution. Specifically, $C \in \mathbb{R}^{d}$, where $d=128$. Thus, the output of the AlexNet, denoted as $X \in \mathbb{R}^D$, where $D=768$, is given as input to the target network, $F(X; \Theta)$, where $\Theta$ denotes the parameters learnt by $H(C;\Phi)$. Formally:

\begin{equation}
    \Theta = H(C; \Phi)
\end{equation}
\begin{equation}
    \hat{y}=F(X; \Theta)
\end{equation}

The hypernetwork consists of fully connected layers and  ReLU activation functions transforming into a meaningful weight representation. In terms of the implementation details, the hypernetwork includes a hidden layer with 512 units and an output layer of 1536 units. The hypernetwork outputs a weight matrix $W \in \mathbb{R}^{768 \times 2}$ and a bias term $b \in \mathbb{R}^2$. The target network performs a matrix multiplication between the extracted features ($X$) and the generated weight matrix. The bias term is added.

\noindent \textbf{Output.} Let $\hat{y} \in \mathbb{R}^2$ denote the output of the target network ($F(X;\Theta)$), since our task corresponds to a binary classification task. We minimize the cross-entropy loss function.

\section{Experiments and Results} \label{experiments_results}

\subsection{Baselines} \label{baselines}

\begin{itemize}
    \item /pa/ + eGeMAPS + Random Forest: This method utilizes the audio files corresponding to the syllable repetition /pa/. Then, this method uses the openSMILE toolkit \cite{10.1145/1873951.1874246} and extracts the eGeMAPS feature set (88d). Then, a Random Forest classifier is trained.
    \item vocalization of the days of the week, reading task, and monologue features + Random Forest: This method utilizes the features provided by the authors in \cite{dubbioso2024voice}, including mean and std F0, jitter, shimmer, etc., and trains an RF classifier.
    \item Introduced approach with inputs the phonations of vowels and syllable repetitions: This method uses as inputs to the proposed deep neural network described in Section~\ref{methodology} speech signals corresponding to the phonation of vowels /a/, /e/, /i/, /o/, /u/, and the syllable repetitions of /ta/ and /ka/.
    \item Multimodal Fusion Method with input /pa/ and /ta/: This method uses as input speech signals corresponding to the /pa/ and /ta/ syllable repetitions. The audio signals are converted into log-Mel spectrogram, delta, and delta-delta and are given as input to pretrained AlexNet models sharing the same weights. The output vectors of the AlexNet models (768d) are fed into a fusion method, namely Gated Multimodal Unit (GMU) \cite{arevalo2020gated}. Let $f^t$ and $f^v$ denote the /pa/ and /ta/ representation vectors respectively. The equations governing the GMU are described as follows:    
    $h^t = \tanh{(W^t f^t + b^t)}, h^v = \tanh{(W^v f^v + b^v)}, z = \sigma(W^z \left [f^v;f^t \right] + b^z), h = z * h^v + (1-z)*h^t, \Theta	= \{W^t, W^v, W^z\}$, where
$\Theta$ denote the learnable parameters, and [.;.] the concatenation operation. $\sigma$ is the sigmoid activation function. $h$ is the output of the GMU. The output of GMU is passed through a dense layer with two units.
    \item Concatenation of all speech signals: This method uses as input the speech signals corresponding to the phonation of the vowels /a/, /e/, /i/, /o/, /u/ and /pa/, /ta/, /ka/ syllable repetition. Firstly, speech signals are converted into log-Mel spectrogram, delta, and delta-delta. Secondly, they are given as input to pretrained AlexNet models sharing the same weights. The output vectors of the AlexNet models (32d) are concatenated into one vector. The resulting vector (256d) is passed through a dense layer of two units.
\end{itemize}

\subsection{Experimental Setup}

We use a 5-fold cross-validation framework with four repetitions to train and test our proposed model. We train the proposed deep learning model for 30 epochs. We use PyTorch for performing our experiments. We use a learning rate of \texttt{1e-5}. Experiments are conducted on a single NVIDIA A100 PCIe 80GB GPU.


\subsection{Evaluation Metrics}

Accuracy, Precision, Recall, and F1-score are used to evaluate our introduced approach. We report the mean and standard deviation of these metrics obtained over 10 runs in a 5-fold cross-validation scheme.

\subsection{Results}

Table~\ref{performance_comparison} reports the results of our proposed approach. Specifically, this table presents a comparison of our approach with the baselines described in Section~\ref{baselines}. As one can observe, our introduced model surpasses the baselines in Recall by 0.83-31.78\%, in F1-score by 2.72-28.95\%, and in Accuracy by 3.02-22.61\%. We observe that the proposed approach with the input of the vowel /u/ yields the highest Specificity accounting for 89.59\%. However, F1-score is a more important metric than Specificity, since a lower F1-score means that dysarthric ALS patients are misdiagnosed as non-dysarthric ones. As one can observe in Table~\ref{performance_comparison}, syllable repetitions of /pa/, /ka/, and /ta/ perform better than the vowels. We also observe that the vowel /u/ presents the best performance among the other vowels. Additionally, the train of a Random Forest classifier utilizing the eGeMAPS feature set of /pa/ yields the lowest evaluation metrics. The fusion of the input representations of /pa/ and /ta/ via a Gated Multimodal Unit presents a decrease in Accuracy in comparison with the unimodal models. Specifically, \textit{Fusion (/pa/+/ta/)} presents a decrease in Accuracy in comparison with /pa/ (5.42\%) and /ta/ (3.02\%). Finally, the fusion method of concatenation yields lower Accuracy and F1-score compared with most vowels or syllables. We hypothesize that these differences in performances are attributable to the fact that concatenation does not capture the inherent correlations of the input representations.

\begin{table}[htbp]
\tiny
\centering
\caption{Performance comparison among proposed models and baselines. Best results per evaluation metric are in bold.}
\begin{tabular}{lccccc}
\toprule
\multicolumn{1}{l}{}&\multicolumn{5}{c}{\textbf{Evaluation metrics}}\\
\cline{2-6} 
\multicolumn{1}{l}{\textbf{Architecture}}&\textbf{Precision}&\textbf{Recall}&\textbf{F1-score}&\textbf{Accuracy}&\textbf{Specificity}\\
\midrule
\multicolumn{6}{>{\columncolor[gray]{.8}}l}{\textbf{Comparison with baselines}} \\
\textit{/pa/+eGeMAPS+RF} & 63.62 & 55.00 & 58.27 & 63.23 & 70.86 \\ 
& $\pm$13.99 & $\pm$16.12 & $\pm$13.78 & $\pm$10.72 & $\pm$12.84 \\ \hline
\makecell[l]{\textit{weekdays,reading},\\ \textit{monologue features+RF}} & 62.45 & 48.44 & 52.46 & 60.05 & 71.09 \\ 
& $\pm$13.50 & $\pm$17.96 & $\pm$12.41 & $\pm$8.52 & $\pm$14.77 \\ \hline
\textit{/ta/} & 80.13 & 79.39 & 78.69 & 79.64 & 79.45 \\ 
& $\pm$9.12 & $\pm$11.34 & $\pm$5.35 & $\pm$4.50 & $\pm$11.63 \\ \hline
\textit{/ka/} & 81.78 & 76.94 & 78.60 & 79.61 & 82.09 \\ 
& $\pm$12.42 & $\pm$6.63 & $\pm$6.54 & $\pm$7.28 & $\pm$14.14 \\ \hline
\textit{/a/} & 82.94 & 67.77 & 72.75 & 76.74 & 85.14 \\ 
& $\pm$11.96 & $\pm$16.16 & $\pm$10.57 & $\pm$6.97 & $\pm$11.79 \\ \hline
\textit{/e/} & 78.76 & 66.72 & 70.98 & 74.43 & 81.45 \\ 
& $\pm$11.16 & $\pm$13.71 & $\pm$8.42 & $\pm$6.21 & $\pm$12.78 \\ \hline
\textit{/i/} & 77.62 & 74.94 & 75.26 & 77.19 & 79.23 \\ 
& $\pm$10.35 & $\pm$16.16 & $\pm$10.77 & $\pm$8.10 & $\pm$10.25 \\ \hline
\textit{/o/} & 80.00 & 72.99 & 75.35 & 77.24 & 81.18 \\ 
& $\pm$9.71 & $\pm$10.63 & $\pm$5.27 & $\pm$4.79 & $\pm$11.76 \\ \hline
\textit{/u/} & \textbf{86.65} & 66.44 & 74.35 & 78.45 & \textbf{89.59} \\ 
& $\pm$9.44 & $\pm$12.12 & $\pm$8.45 & $\pm$6.43 & $\pm$8.47 \\ \hline
\textit{Fusion (/pa/+/ta/)} & 76.96 & 75.66 & 75.70 & 77.21 & 78.77 \\ 
& $\pm$10.70 & $\pm$14.82 & $\pm$11.06 & $\pm$9.69 & $\pm$10.25 \\ \hline
\textit{Concatenation} & 77.75 & 72.94 & 74.40 & 76.05 & 78.82 \\ 
& $\pm$12.39 & $\pm$13.26 & $\pm$9.93 & $\pm$9.31 & $\pm$14.90 \\ 

\midrule
\multicolumn{6}{>{\columncolor[gray]{.8}}l}{\textbf{Introduced Approach}} \\
\textit{/pa/} & 84.18 & \textbf{80.22} & \textbf{81.41} & \textbf{82.66} & 84.73 \\ 
  & $\pm$7.99 & $\pm$11.15 & $\pm$6.10 & $\pm$5.44 & $\pm$9.34 \\ 

\bottomrule
\end{tabular}
\label{performance_comparison}
\end{table}

\subsection{Ablation Study}

In this section, we perform a series of ablation experiments to investigate the effectiveness of the proposed approach. Firstly, we replace the hypernetwork with a simple dense layer. Therefore, the output vector of the AlexNet model is fed to a dense layer of two units. Findings showed that Accuracy and F1-score presented a decrease of 2.46\% and 2.08\% respectively. Secondly, we use a data-conditioned hypernetwork. Specifically, we use the eGeMAPS feature set (88d) as the condition vector, i.e., input to the hypernetwork $H(C; \Phi)$, instead of a vector following a normal distribution. Results showed that Accuracy and F1-score presented a decline of 4.22\% and 4.98\% respectively. Thirdly, we use as input MFCC instead of log-Mel spectrogram features. Therefore, the input to the deep neural network is the transformation of the speech signal into MFCC, delta, and delta-delta. Results indicated that Accuracy and F1-score dropped by 2.48\% and 3.69\% respectively. Finally, we do not use pretrained AlexNet model. Specifically, we use AlexNet with no pretrained weights. Findings showed that Accuracy and F1-score had a decrease of 3.33\% and 4.53\% respectively.

\begin{table}[htbp]
\tiny
\centering
\caption{Ablation Study. Best results per evaluation metric are in bold.}
\begin{tabular}{lccccc}
\toprule
\multicolumn{1}{l}{}&\multicolumn{5}{c}{\textbf{Evaluation metrics}}\\
\cline{2-6} 
\multicolumn{1}{l}{\textbf{Architecture}}&\textbf{Precision}&\textbf{Recall}&\textbf{F1-score}&\textbf{Accuracy}&\textbf{Specificity}\\
\midrule
\multicolumn{6}{>{\columncolor[gray]{.8}}l}{\textbf{Ablation Experiments}} \\
\textit{Removal of Hypernetwork} & 81.25 & 78.66 & 79.33 & 80.20 & 81.68\\ 
& $\pm$11.66 & $\pm$9.31 & $\pm$7.62 & $\pm$7.88 & $\pm$12.18 \\ \hline
\textit{Usage of eGeMAPS as condition vector} & 80.07 & 74.88 & 76.43 & 78.44 & 81.68 \\ 
& $\pm$10.86 & $\pm$13.45 & $\pm$9.29 & $\pm$7.68 & $\pm$10.73 \\ \hline
\textit{MFCC} & 85.48 & 73.50 & 77.72 & 80.18 & 86.14 \\ 
& $\pm$10.75 & $\pm$13.50 & $\pm$8.26 & $\pm$6.79 & $\pm$12.10 \\ \hline
\textit{AlexNet with no pretrained weights} & 82.48 & 73.94 & 76.88 & 79.33 & 84.50\\ 
& $\pm$10.00 & $\pm$14.88 & $\pm$9.52 & $\pm$7.69 & $\pm$9.02 \\ 
\midrule
\multicolumn{6}{>{\columncolor[gray]{.8}}l}{\textbf{Introduced Approach}} \\
& \textbf{84.18} & \textbf{80.22} & \textbf{81.41} & \textbf{82.66} & \textbf{84.73} \\ 
& $\pm$7.99 & $\pm$11.15 & $\pm$6.10 & $\pm$5.44 & $\pm$9.34 \\
\bottomrule
\end{tabular}
\label{ablation_study}
\end{table}

\section{Conclusion} \label{conclusion_limitations_futurework}

In this paper, we present the first study integrating hypernetworks into a deep neural network for identifying dysarthria in ALS patients. Specifically, after converting each audio signal into an image of three channels, namely log-Mel spectrogram, delta, and delta-delta, we pass the resulting image into AlexNet. Then, a hypernetwork is used for producing weights for the target network. Specifically, the output vector of AlexNet is given as input to the target network, while a vector with normal distribution is given as input to the hypernetwork. Results showed that the proposed approach yielded an Accuracy of 82.66\%, while results of an ablation study demonstrated the effectiveness of the introduced approach. \textbf{Limitations:} The VOC-ALS dataset is imbalanced in terms of the severity levels of dysarthria in ALS patients. For this reason, in this study, we did not experiment with predicting the severity level of dysarthria. \textbf{Future Work:} In the future, we aim to use neural architecture search approaches for finding the optimal architecture in our task.

\bibliographystyle{IEEEtran}
\bibliography{mybib}

\end{document}